# Fourier Imager Network (FIN): A deep neural network for hologram reconstruction with superior external generalization

Hanlong Chen[1,2,3]†, Luzhe Huang[1,2,3]†, Tairan Liu[1,2,3], Aydogan Ozcan[1,2,3,4]*

1 Electrical and Computer Engineering Department, University of California, Los Angeles, California 90095, USA

2 Bioengineering Department, University of California, Los Angeles, California 90095, USA

3 California Nano Systems Institute (CNSI), University of California, Los Angeles, California 90095, USA

4 David Geffen School of Medicine, University of California Los Angeles, California 90095, USA

† Contributed equally

Email addresses:

hanlong@ucla.edu, lzhuang0324@ucla.edu, liutr@ucla.edu, ozcan@ucla.edu

* Corresponding author: Aydogan Ozcan, ozcan@ucla.edu

Address: 68-119 Engr. IV, 420 Westwood Plaza, UCLA, Los Angeles, CA 90095, USA

# Abstract

Deep learning-based image reconstruction methods have achieved remarkable success in phase recovery and holographic imaging. However, the generalization of their image reconstruction performance to new types of samples never seen by the network remains a challenge. Here we introduce a deep learning framework, termed Fourier Imager Network (FIN), that can perform end-to-end phase recovery and image reconstruction from raw holograms of new types of samples, exhibiting unprecedented success in external generalization. FIN architecture is based on spatial Fourier transform modules that process the spatial frequencies of its inputs using learnable filters and a global receptive field. Compared with existing convolutional deep neural networks used for hologram reconstruction, FIN exhibits superior generalization to new types of samples, while also being much faster in its image inference speed, completing the hologram reconstruction task in ~0.04 s per 1 $mm^2$ of the sample area. We experimentally validated the performance of FIN by training it using human lung tissue samples and blindly testing it on human prostate, salivary gland tissue and Pap smear samples, proving its superior external generalization and image reconstruction speed. Beyond holographic microscopy and quantitative phase imaging, FIN and the underlying neural network architecture might open up various new opportunities to design broadly generalizable deep learning models in computational imaging and machine vision fields.

# Introduction

Digital holography provides unique advantages in microscopic imaging, by reconstructing the complex optical fields of input samples[1–12]. Due to the missing phase information, various computational approaches have been developed to digitally reconstruct holograms[13–22]. Recent work has also utilized deep neural networks[23–43] to reconstruct the complex sample field from a hologram in a single forward inference step, achieving an image reconstruction quality comparable to iterative hologram reconstruction algorithms that are based on physical wave propagation. Some of the earlier results have also reported simultaneous performance of phase retrieval and autofocusing in a single network architecture, demonstrating holographic imaging over an extended depth-of-field[25,35,42]. In these earlier demonstrations, various deep network architectures, such as e.g., U-net-based convolutional neural networks (CNNs),[23,25,27,34,38] recurrent neural networks (RNNs)[42,44], as well as generative adversarial networks (GANs)[32,38,42,43,45] have been proven to be effective for phase retrieval and hologram reconstruction for new (unseen) objects that belong to the same sample type of interest used during the training process. Stated differently, this earlier body of work has successfully demonstrated the "internal generalization" of the hologram reconstruction and phase retrieval networks to new objects of the same sample type as used in training.

On the other hand, "external generalization" to new objects from entirely new types of samples, never seen by the network before, remains a major challenge for deep neural networks, which might lead to image reconstruction degradation or hallucinations. Some studies have explored using transfer learning to address this challenge, which requires fine-tuning the network using a subset of the new types of samples[46–48]. In addition to this external generalization issue, it is, in general, difficult for CNN-based image reconstruction networks to accurately reconstruct raw holograms of samples due to the limited receptive field of convolutional layers, which casts another challenge considering the relatively large scale of holographic diffraction patterns of samples. As a result, existing end-to-end hologram reconstruction deep neural networks[24,26,34,36,42] could only achieve decent reconstruction performance on relatively sparse samples. Alternatively, a pre-processing step, such as zero phase-padded free space propagation (FSP), has also been utilized to better deal with this issue[23,25,26,41,42], which requires a precise physical forward model with the correct estimate of the axial propagation distance.

Here we introduce an end-to-end deep neural network, termed Fourier Imager Network (FIN), to rapidly implement phase recovery and holographic image reconstruction from raw holograms of new types of samples, achieving unprecedented success in external generalization. This framework takes in two or more input raw holograms captured at different sample-to-sensor distances without any pre-processing steps involved. By comprehensively utilizing the global spatial-frequency information processed by its trained spatial Fourier transform (SPAF) modules, FIN accurately reconstructs the complex field of the specimen, successfully demonstrating external generalization to new types of samples never used during its training. To experimentally demonstrate the success of this approach, we trained FIN models using human lung tissue samples (i.e., thin histopathology sections of connected tissue) and blindly tested the resulting trained FIN models on prostate and salivary gland tissue sections as well as Pap smear samples without compromising the image reconstruction quality. Compared to iterative hologram reconstruction algorithms based on wave propagation between different measurement planes, FIN is >27-fold faster to reconstruct an image. Compared to existing CNN-based deep learning models, FIN exhibits an unprecedented generalization performance, and is also much faster in its inference speed. We expect FIN to be widely used in various image reconstruction and enhancement tasks commonly employed in the computational microscopy field. In addition to coherent imaging, FIN can be applied to other image reconstruction or enhancement tasks in different imaging modalities, including e.g., fluorescence and brightfield microscopy.

## Results

FIN provides an end-to-end solution for phase recovery and holographic image reconstruction, and its architecture is schematically presented in Fig. 1 (also see the Methods section). To acquire raw holograms of specimens, we used a lens-free in-line holographic microscope, as detailed in the Methods section, to image transmissive samples, such as human tissue samples and Pap smears, using a set of sample-to-sensor distances, i.e., $z_{2,i}, i = 1, \cdots, M$. The input images to FIN consist of $M$ intensity-only raw holograms captured at $z_{2,1}$ to $z_{2,M}$ and the network outputs are the reconstructed real and imaginary images of the object, revealing the complex-valued sample field. The corresponding ground truth images for supervised learning of FIN are obtained using an iterative multi-height phase retrieval (MH-PR) algorithm[18] with $M = 8$ holograms acquired at different sample-to-sensor distances.

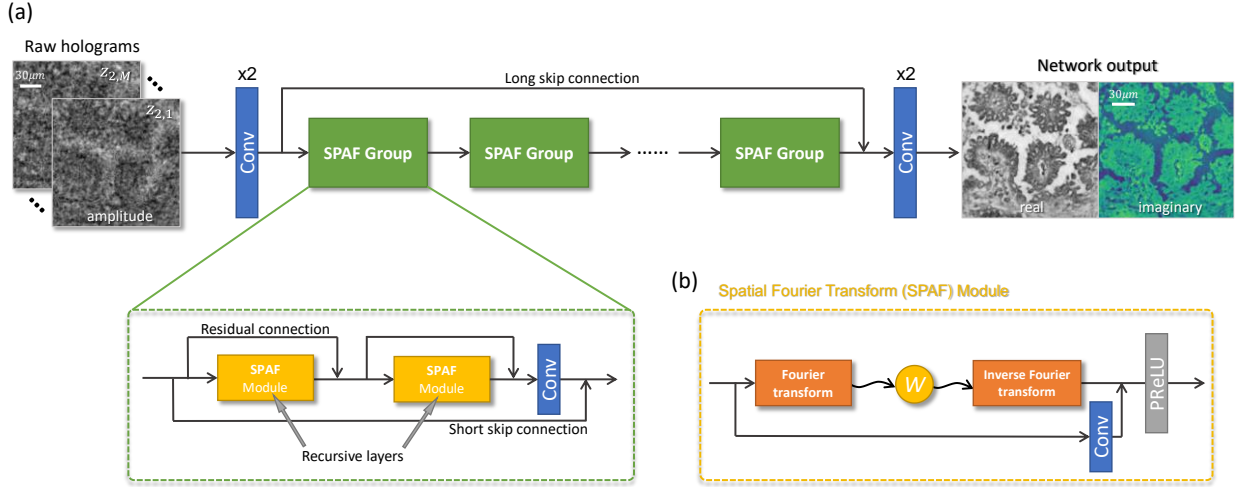

**Figure 1** (a) The model architecture of FIN. The raw input holograms are captured at different sample-to-sensor distances $z_{2,i}$, $i = 1, \ldots, M$. (b) The structure of the SPAF Module, where $W$ stands for the element-wise multiplication in the frequency domain (see the Methods section for details).

To demonstrate the success of FIN, we trained it using raw holograms of human lung tissue sections and tested the trained model on four different types of samples: (1) lung tissue samples from different patients never used in the training set (testing *internal* generalization), (2) Pap smear samples, (3) prostate tissue samples, and (4) salivary gland tissue samples, where (2,3,4) test *external* generalization, referring to new types of samples. The $z_{2,i}$ distances that we used in these holographic imaging experiments were 300, 450, and 600 $\mu m$ ($M = 3$). After its training, our blind testing results (Fig. 2) reveal that FIN can not only reconstruct new lung tissue sections from new patients (internal generalization) but also achieves a strong external generalization performance on new sample types never seen by the network before. Furthermore, compared to the output of the MH-PR algorithm using the same input (raw hologram) data ($M = 3$), FIN is ~27.3 times faster in its inference speed per image (see Table 1) and delivers better reconstruction quality on all the test samples, as highlighted by the yellow arrows and the zoomed-in regions in Fig. 2.

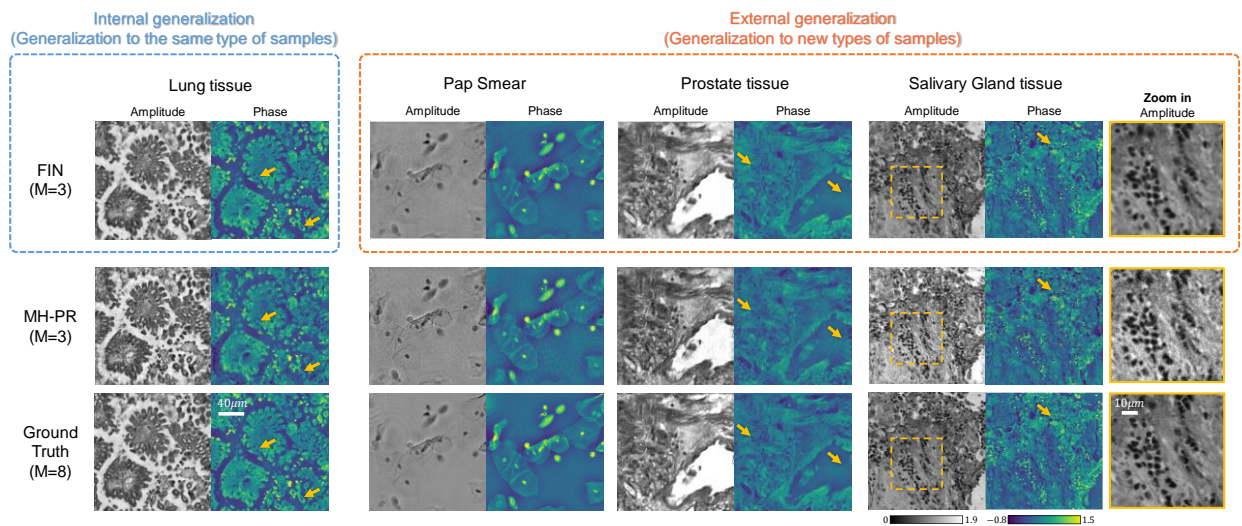

**Figure 2** Internal and external generalization of FIN. Each output of FIN and MH-PR algorithm is generated using the same raw holograms (M=3). FIN was trained on human lung tissue samples only, and the internal generalization part uses unseen lung tissue holograms from new patients. The external generalization directly applies the same trained FIN model to new types of samples, i.e., Pap smear, prostate tissue, and salivary gland tissue samples, never seen by the network before. The ground truth for each sample is obtained through the MH-PR algorithm that used M=8 raw holograms captured at different sample-to-sensor distances.

To further showcase the generalization ability of FIN, we separately trained four FIN models using lung tissue, prostate tissue, salivary gland tissue, and Pap smear hologram datasets (i.e., one type of sample for each network model), and blindly tested the trained models on unseen FOVs from four types of samples using M=3 raw holograms for each FOV. Similar to our conclusions reported in Fig.2, Fig. 3(a) also shows that the FIN model trained using each sample type can successfully generalize to other types of samples. Even when FIN was only trained using relatively sparse samples such as Pap smear slides, the resulting network model successfully generalized to reconstruct the raw holograms of connected tissue sections that significantly deviate from the structural distribution and sparsity observed in Pap smear samples. The success of the reconstruction performance of FIN was also quantified using the structural similarity index (SSIM)[49] across all four FIN networks that were trained using different types of samples (see Fig. 3(b)), demonstrating the superior generalization of these FIN models regardless of the distribution shifts observed between the training and testing data.

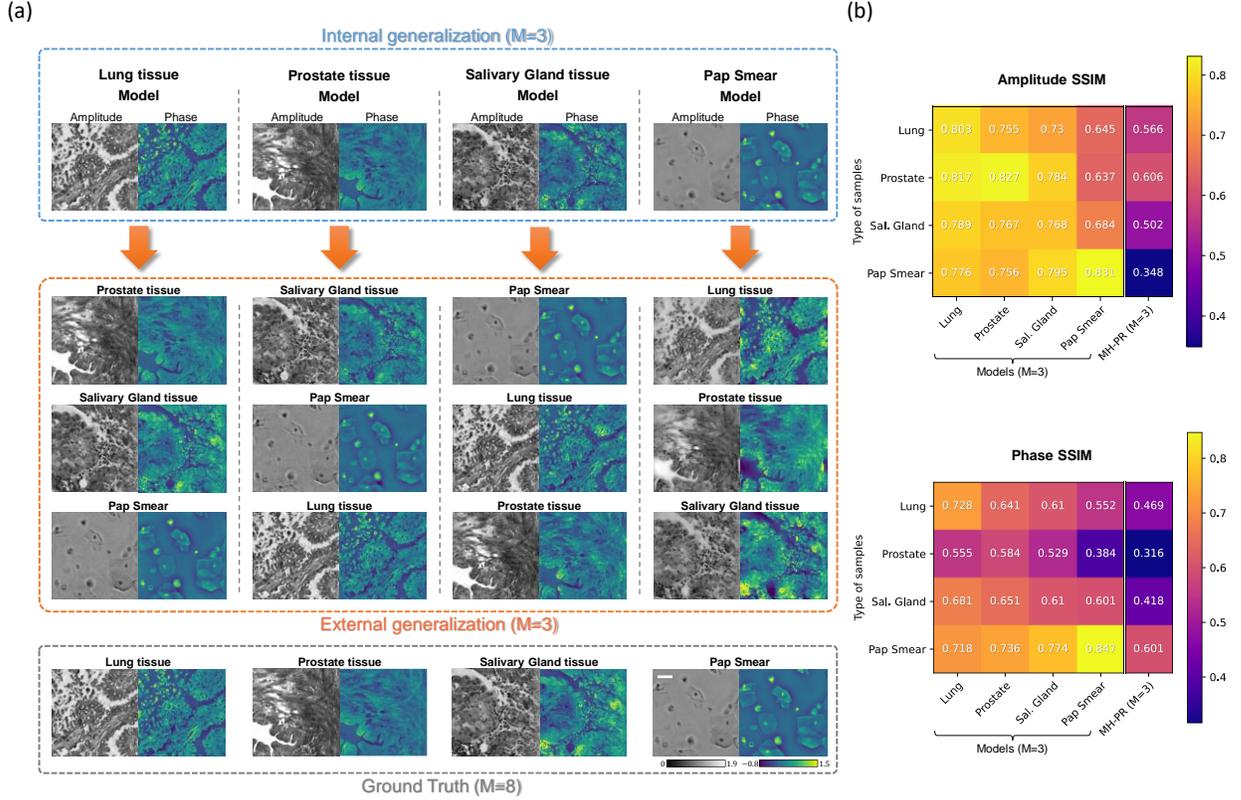

**Figure 3** Generalization of FIN on different types of training samples. (a) Each FIN model is separately trained using the raw holograms (M=3) of the corresponding sample type. Scale bar: 30 $\mu m$. (b) A comparison of the amplitude and phase SSIM values corresponding to the hologram reconstruction performed by different FIN models that are trained and tested with different types of samples. The ground truth for each sample is obtained through the MH-PR algorithm that used M=8 raw holograms captured at different sample-to-sensor distances.

Next, we evaluated the hologram reconstruction performance of FIN when only two input holograms were measured, i.e., M=2. For this, we trained ten different FIN models from scratch using the same human lung tissue sections but with different sets of $z_{2,1}$ and $z_{2,2}$, such that the sample-to-sensor distances for different FIN models were different. These trained FIN models were then blindly tested on new lung tissue sections from new patients (internal generalization); Fig. 4 reports the amplitude and phase root mean square error (RMSE) of the reconstructed holographic images generated by FIN (M=2) and MH-PR (M=2) for different combinations of $z_{2,1}$ and $z_{2,2}$. Both the amplitude and phase RMSE values show that FIN achieves a significant reconstruction quality improvement compared to MH-PR (M=2), and the RMSE values of FIN models are consistently better with different sample-to-sensor distances varying from 300 $\mu m$ to 600 $\mu m$. The visualization of the reconstructed holograms shown in Fig. 4 further confirms the same conclusion that FIN achieves consistently better image reconstruction compared to MH-PR for various combinations of $z_{2,1}$ and $z_{2,2}$. We also confirmed that the same conclusions apply to the external generalization tests of FIN (M=2).

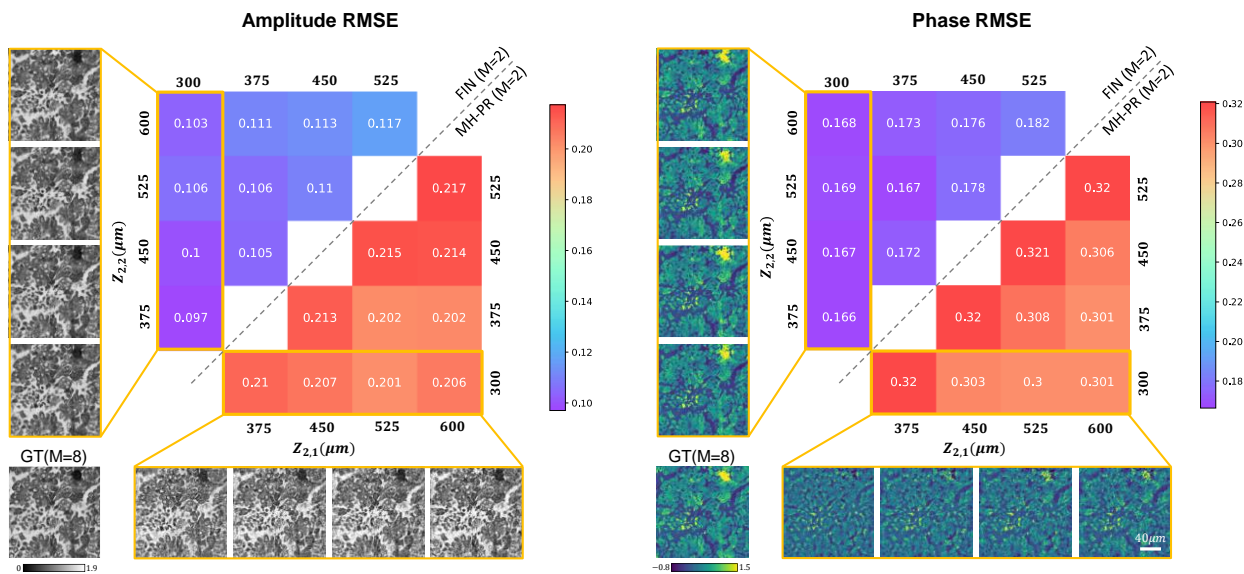

**Figure 4** A comparison of amplitude and phase RMSE values using different combinations of $z_{2,1}$ and $z_{2,2}$. Each number in the FIN region refers to an independent network model trained using the corresponding $z_{2,1}$ and $z_{2,2}$ combination (M=2). The ground truth for each sample is obtained through the MH-PR algorithm that used M=8 raw holograms captured at different sample-to-sensor distances. GT: Ground Truth.

In addition to MH-PR based comparisons, we also extended our performance analysis to other deep learning-based phase retrieval and hologram reconstruction methods. For this additional set of comparisons, we used a state-of-the-art deep learning model based on a recurrent convolutional neural network, termed RH-M, that was developed for multi-height holographic image reconstruction.[42] Using the same training hologram data, we trained FIN and RH-M models for different M values, the blind testing results of which are compared in Fig. 5. As for the internal generalization performance shown in Fig. 5, both FIN and RH-M can successfully generalize to new lung tissue samples from new patients. However, for the external generalization to new sample types (prostate and salivary gland tissue as well as Pap smear samples), FIN provides superior image reconstruction performance even though it uses a smaller number of trainable parameters compared to RH-M (see Table 1); in comparison, RH-M has reconstruction artifacts on external testing sets for both M=3 and M=4, also confirmed by the significantly lower SSIM values (for RH-M reconstructions) reported in Fig. 5.

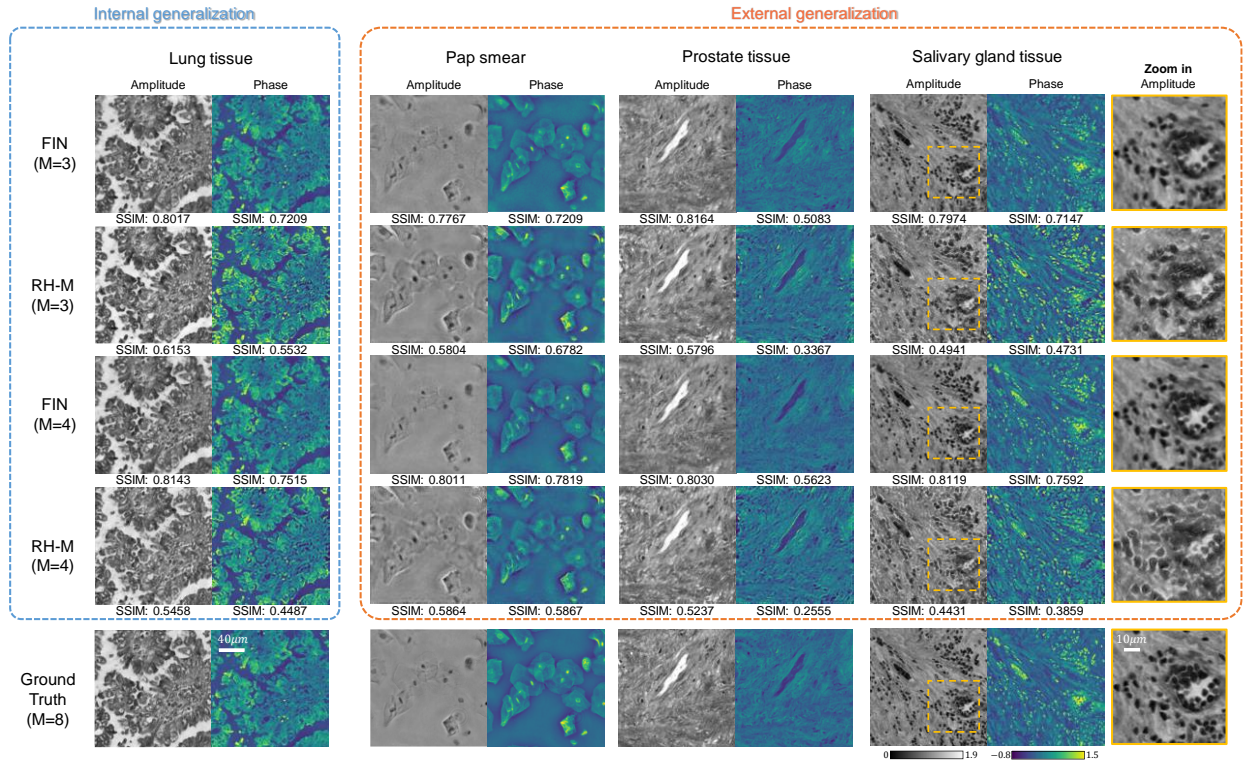

**Figure 5** A comparison of the generalization capabilities of FIN and RH-M. Both FIN and RH-M are trained using only human lung tissue samples and the internal generalization part uses unseen FOVs from new lung tissue samples (new patients). The external generalization tests directly apply the same trained networks to new types of samples, i.e., Pap Smear, prostate tissue and salivary gland tissue samples. The ground truth for each sample is obtained through the MH-PR algorithm that used M=8 raw holograms captured at different sample-to-sensor distances.

In addition to its superior generalization performance, FIN also has faster inference speed compared to deep learning-based or iterative phase retrieval algorithms. In Table 1, we compared the inference time of FIN, RH-M, and MH-PR algorithms. Noticeably, FIN has the shortest inference time among these methods using any number of raw input holograms. For the case of M=3, FIN is ~9.3-fold faster than RH-M and ~27.3-fold faster than MH-PR, which highlights the computational efficiency of our network. We can further accelerate the inference speed of FIN by using parallelization, which reduces the computation time to $0.04\ s/mm^2$ under an image batch size of 20 (see Table 1). We should also note that the number (M) of input holograms has a negligible impact on the inference time of FIN, since it uses a fixed channel size for most parts of the network model, and M only affects the first 1x1 convolutional layer. That is why the inference times of FIN (M=3) and FIN (M=4) are approximately the same as shown in Table 1. Refer to the Methods section for further details.

**Table 1. Inference Time Comparison of FIN, RH-M, and MH-PR Algorithms.**

|  | Number of trainable parameters | Inference time ($s/mm^2$) | Parallelized inference time ($s/mm^2$) |
|---|---|---|---|
| FIN (M=3) | 11.5M | 0.52 | 0.04 |
| FIN (M=4) | 11.5M | 0.56 | 0.04 |
| RH-M (M=3) | 14.1M | 4.84 | 1.96 |
| RH-M (M=4) | 14.1M | 5.76 | 2.08 |
| MH-PR (M=2) | N/A | 10.36 | N/A |
| MH-PR (M=3) | N/A | 14.19 | N/A |

The parallelized inference time is measured when the batch size is set to 20 for both FIN and RH-M.

# Discussion

We demonstrated an end-to-end phase retrieval and hologram reconstruction network that is highly generalizable to new sample types. FIN outperforms other phase retrieval algorithms in terms of both the reconstruction quality and speed. This method presents superior generalization capability to new types of samples without any prior knowledge about these samples or any fine-tuning of its trained model. This strong external generalization of our model mainly stems from the regularization effect of the SPAF modules in its architecture. In a lensfree holographic imaging system, the Fourier transforms of the fields at the sample plane and the measurement plane are related by a frequency-dependent phasor, which can be effectively learned through the element-wise multiplication module in SPAF. Besides, SPAF modules provide a global receptive field to FIN, in contrast to the limited, local receptive fields of common CNNs. The global receptive field helps the FIN model more effectively process the holographic diffraction patterns for various samples, regardless of the morphologies and dimensions of the objects. In fact, previous research has already shown that end-to-end hologram reconstruction requires a larger network receptive field, which can be partially addressed by using e.g., dilated convolution[42]. In our method, the Fourier transform intrinsically captures the global spatial information of the sample and thus provides a maximized receptive field for FIN, contributing to its performance gain over CNN-based hologram reconstruction models reported in Fig. 5. Like FIN, other deep neural networks[50] have also utilized learnable spatial Fourier transform modules for inference, for example, to successfully map the initial and/or boundary conditions of partial differential equations (PDEs) and infer numerical solutions.

Unlike fully convolutional networks, in FIN architecture, the size of the input raw hologram FOV is fixed at the beginning, i.e., we cannot use a larger FOV in the testing phase because of the element-wise multiplication in our SPAF module. A larger FOV raw hologram can be reconstructed using FIN by dividing the hologram into smaller FOVs and running them through FIN in parallel. This parallelization of a large FOV hologram reconstruction is feasible since FIN has a significant speed advantage in its inference, and can reconstruct ~1 mm$^2$ sample area within 0.04 sec using a standard GPU (see Table 1).

## Methods

**Holographic Imaging:** A lens-free in-line holographic microscope was utilized to capture the raw holograms of the specimens. A broadband light source (WhiteLase Micro, NKT Photonics) and an acousto-optic tunable filter (AOTF) were used as the illumination source emitting 530 nm light. The image sensor was a complementary metal-oxide-semiconductor (CMOS) RGB image sensor (IMX081, Sony). The light source, sample, and the CMOS image sensor were aligned vertically. The sample was directly placed between the light source and the sensor such that the sample-to-source distance ($z_1$) was about 10 cm, and the sample-to-sensor distance ($z_2$) ranged from 300 $\mu m$ to 600 $\mu m$. The CMOS sensor was placed on and controlled by a 6-axis stage (MAX606, Thorlabs) to perform lateral and axial shifts. All hardware was connected to a computer and controlled by a customized LabVIEW program to capture holograms automatically.

All the human samples involved in this work were deidentified and prepared from existing specimens that were captured before this research. Human prostate, salivary gland, and lung tissue slides were provided by the UCLA Translational Pathology Core Laboratory (TPCL). Pap smear slides were prepared by the UCLA Department of Pathology.

**Pre-processing:** The captured raw holograms were firstly processed by a pixel super-resolution algorithm[51,18,52]. The 6-axis stage was programmed to automatically capture in-line holograms at 6x6 lateral positions with sub-pixel shifts. The super-resolution algorithm estimated the relative shifts for each hologram and merged these holograms using a shift-and-add algorithm[18]. The effective pixel size of the generated super-resolved holograms decreases to 0.37 $\mu m$ from the original CMOS pixel size of 2.24 $\mu m$. The resulting super-resolved holograms were cropped into unique patches of 512x512 pixels, without any overlap. Hologram datasets of each sample type were partitioned into training and testing sets, at a ratio of 6:1, comprising ~600 unique FOVs in each training set and ~100 FOVs for the testing set. The testing FOVs were strictly obtained from different whole slides (new patients) excluded in the training sets.

The ground truth sample fields were retrieved by an iterative multi-height phase retrieval algorithm[18]. At each sample FOV, M=8 in-line holograms were captured at different sample-to-sensor distances, which were later estimated by an autofocusing algorithm using the edge sparsity criterion[53]. In each iteration, the estimated sample field is digitally propagated to each hologram plane using the angular spectrum propagation[54]. The propagated complex field is updated according to the measurement at each hologram plane, by averaging the amplitude of the propagated field with the measured amplitude and retaining the new estimated phase. One iteration is completed after all the hologram planes are used, and this MH-PR algorithm converges within 100 iterations.

**Network structure:** The FIN network architecture has a Residual in Residual architecture shown in Fig. 1, inspired by RCAN[55] to have a deeper network structure and better information flow. Our network, FIN, consists of several SPAF modules with a long skip connection to form the large-scale residual connection, in conjunction with two 1x1 convolutional layers at the head and tail of the network. Each SPAF group contains two recursive SPAF modules, which share the same parameters to improve the network capacity without significantly enlarging the size of the network. A short skip connection is introduced for every SPAF group to form the middle-scale residual connection, and a small-scale residual connection is used to connect the inputs and outputs of each SPAF module. SPAF module, as shown in Fig. 1(b), has a linear transformation applied to the tensor after it was transformed into the frequency domain using the 2D Discrete Fourier transform, following a similar architecture as in Ref.[50]; a half window size of $k$ is applied to truncate the higher frequency signals, i.e.,

$$F'_{j,u,v} = \sum_{i=1}^{c} W_{i,j,u,v} \cdot F_{i,u,v}, \quad u,v = 0, \pm 1, \dots, \pm k, \quad j = 1, \dots, c$$

where $F \in \mathbb{C}^{c,2k+1,2k+1}$ is the truncated frequency domain representation of the input to the SFAP module after performing the 2D Discrete Fourier Transform, $W \in \mathbb{R}^{c,c,2k+1,2k+1}$ represents the trainable weights, $c$ is the channel number, and $k$ is the half window size. After this linear transformation, the inverse 2D Discrete Fourier transform is used to obtain the processed data back in the spatial domain, followed by a PReLU activation function.

$$\text{PReLU}(x) = \begin{cases} x, & \text{if } x \geq 0 \\ ax, & \text{otherwise} \end{cases}$$

where $a$ is a learnable parameter.

To adapt the SPAF module to high-resolution image processing in a deeper network, we shrank the matrix $W$ allowing a significant model size reduction. The optimized linear transformation is defined as

$$F'_{j,u,v} = W'_{j,u,v} \cdot \sum_{i=1}^{c} F_{i,u,v}, \quad u,v = 0, \pm 1, \dots, \pm k, \quad j = 1, \dots, c$$

where $F \in \mathbb{C}^{c,2k+1,2k+1}$ is the truncated frequency components, and $W' \in \mathbb{R}^{c,2k+1,2k+1}$ represents the trainable weights.

To further optimize the network structure for high-resolution holographic image reconstruction, a set of decreasing half window sizes $(k)$ was chosen for the SPAF modules. Specifically, both of the SPAF modules in each SPAF group have shared hyperparameters, and we set a decreasing half window size $k$ for the SPAF groups in the sequence of the network structure, which forms a pyramid-like structure. This pyramid-like structure provides a mapping of the high-frequency information of the holographic diffraction patterns to low-frequency regions in the first few layers and passes this low-frequency information to the subsequent layers with a smaller window size, which better utilizes the features at multiple scales and at the same time considerably reduces the model size, avoiding potential overfitting and generalization issues.

**Network implementation** The networks are implemented using PyTorch[56] with GPU acceleration and are trained and tested on the same computer with an Intel Xeon W-2195 CPU, 256GB memory, and NVidia RTX 2080 Ti GPUs. During the training phase, the input FOVs of 512x512 pixels were randomly selected from the training hologram dataset, and data augmentation was applied to each FOV, which includes random image rotations of 0, 90, 180, or 270 degrees.

The training loss is the weighted sum of three different loss terms:
$$L_{loss} = \alpha L_{MAE} + \beta L_{complex} + \gamma L_{percep}$$
where $\alpha$, $\beta$, and $\gamma$ are set as 0.5, 1, and 0.5, respectively. The MAE loss and complex domain loss can be expressed as:
$$L_{MAE} = \frac{\sum_{i=1}^{n}|y_i - \hat{y}_i|}{n}$$
$$L_{complex} = \frac{\sum_{i=1}^{n}|\mathcal{F}(y) - \mathcal{F}(\hat{y})|}{n}$$
where $y$ is the ground truth, $\hat{y}$ is the network's output, $n$ is the total number of pixels, and $\mathcal{F}$ stands for the 2D Discrete Fourier Transform operation. For the perceptual loss term[57], we used a pre-trained VGG16 network as the feature extractor to minimize the Euclidean distance between the low-level features of the reconstructed images and the ground truth images.

The trainable parameters of the deep neural network models are learned iteratively using the Adam optimizer[58] and the cosine annealing scheduler with warm restarts[59] is used to dynamically adjust the learning rate during the training phase.

In the testing phase, a batch of test holograms with the same resolution (512x512 pixels) is fed to the network, and the inference time for one FOV at a time (batch size is set to 1) is 0.52 $s/mm^2$. Additionally, using the same Nvidia RTX 2080 Ti GPU, the inference can be parallelized with a batch size of 20, resulting in 0.04 $s/mm^2$ inference time (Table 1).